\pdfoutput=1

\documentclass[11pt]{article}
\usepackage{graphicx}

\usepackage{ACL2023}
\usepackage{multirow}
\usepackage{times}
\usepackage{latexsym}
\usepackage{adjustbox}
\DeclareUnicodeCharacter{0002}{ }
\usepackage[T1]{fontenc}

\usepackage[utf8]{inputenc}

\usepackage{microtype}

\usepackage{inconsolata}

\title{APE-then-QE: Correcting then Filtering Pseudo Parallel Corpora for MT Training Data Creation}

\author{Akshay Batheja, Sourabh Dattatray Deoghare, Diptesh Kanojia, Pushpak Bhattacharyya\\[1ex]
CFILT, Indian Institute of Technology Bombay \\
\texttt{\{akshaybatheja, pb\}@cse.iitb.ac.in}
}

\begin{document}
\maketitle
\begin{abstract}
Automatic Post-Editing (APE) is the task of automatically identifying and correcting errors in the Machine Translation (MT) outputs. We propose a method that uses an APE system to correct errors on the target side of the MT training data. We select the sentence pairs from the original and corrected sentence pairs based on the quality scores computed using a Quality Estimation (QE) model. To the best of our knowledge, this is a novel adaptation of APE and QE to extract quality parallel corpus from the pseudo-parallel corpus. By training with this filtered corpus, we observe an improvement in the Machine Translation system's performance by \textbf{5.64} and \textbf{9.91} BLEU points, for English-Marathi and Marathi-English language pairs, over the baseline model. The baseline model is the one that is trained on the whole pseudo-parallel corpus. \textit{Our work is not limited by the characteristics of English or Marathi languages; and is language pair-agnostic, given the necessary QE and APE data.}

\end{abstract}

\section{Introduction}
In recent times, Neural MT has shown excellent performance, having been trained on a large amount of parallel corpora~\cite{dabre}. However, not all language pairs have a substantial amount of parallel data. Hence, we have to rely on the noisy web-crawled corpora for low-resource languages. The task of \textbf{Parallel Corpus Filtering} aims to provide a scoring mechanism that helps extract good-quality parallel corpus from a noisy pseudo-parallel corpus. The task of \textbf{Automatic Post Editing} (APE) task aims to identify and correct errors in MT outputs automatically. The task of \textbf{Quality Estimation} (QE) aims to provide a quality score for a translation when the reference translation is unavailable. Our motivation is as follows: As the APE systems have yet to be investigated for error correction in noisy pseudo-parallel corpora, this encourages us to utilize them to enhance the quality of the pseudo-parallel corpus. We use the APE model to rectify errors in the target side of the noisy pseudo-parallel corpus. Next, we employ Quality Estimation to assign quality scores to the sentence pairs present in the original and corrected pseudo-parallel corpus and select the parallel sentences with the highest quality scores. We aim to improve the quality of Machine Translation for the English (En) - Marathi (Mr) language pair by  using an APE model to rectify errors on the target side of the pseudo-parallel corpus. We observe that APE and QE-assisted corpus filtering significantly improves the performance of the En-Mr MT system. Our contributions are:
\begin{enumerate}

    \item \textbf{Adaptation of APE+QE} (i) to correct errors on the target side of the pseudo-parallel corpus using Automatic Post-Editing and (ii) to select the best sentence pairs from the corrected and original pseudo-parallel corpus using Quality Estimation;  to the best of our knowledge, this is a novel adaptation of APE and QE for extracting quality parallel corpus from the pseudo-parallel corpus.
    
    \item Demonstration of \textbf{performance improvement} of the Machine Translation systems by \textbf{5.64} and \textbf{9.91} BLEU points for English-Marathi and Marathi-English language pairs, over the model trained on the whole pseudo-parallel corpus; the choice of language pairs was dictated by the availability of APE and QE data, manually created for English-Marathi pairs.
    
\end{enumerate}


\section{Related work}
\subsection{Parallel Corpus Filtering}
Neural Machine Translation (NMT) is extremely \textit{data hungry}~\cite{sutskever2014sequence, bahadanau, NIPS2017_3f5ee243}. The Conference on Machine Translation (WMT) has organized annual Shared Tasks on Parallel Corpus Filtering (WMT 2018, WMT 2019, WMT 2020).~\citet{labse} proposed the LaBSE model, which is a multilingual sentence embedding model trained on 109 languages, including some Indic languages. \\ 
\indent Most recently,~\citet{akshay} used a combination of Phrase Pair Injection and LaBSE~\cite{labse} based Corpus Filtering to extract high-quality parallel data from a noisy parallel corpus.~\citet{briakou-etal-2022-bitextedit} proposed an approach to reconstruct the original translations and translate, in a multi-task fashion, and use it to automatically edit the mined corpus.  In contrast, we use APE and QE-assisted filtering to correct and extract high-quality data from noisy pseudo-parallel data.  

\subsection{Quality Estimation}
Quality Estimation (QE) is the task of evaluating the quality of a translation when reference translation is not available. The state-of-the-art MonoTransQuest architecture, proposed by~\citet{ranasinghe-etal-2020-transquest}, builds upon XLM-R, a widely-used pre-trained cross-lingual language model known for its ability to generalize to low-resource languages~\cite{conneau-etal-2020-unsupervised}.~\citet{kocyigit-etal-2022-better} proposed a combination of multitask training, data augmentation
and contrastive learning to achieve better and
more robust QE, in Parallel Corpus Mining setting. 

\subsection{Automatic Post Editing}
Automatic Post-Editing (APE) is an auxiliary task in a Machine Translation (MT) field, which is aimed at automatically identifying and correcting MT output errors~\cite{chatterjee-etal-2020-findings}. APE systems have the potential to reduce human effort by correcting systematic and repetitive translation errors~\cite{laubli-etal-2013-assessing, pal-etal-2016-multi}. Recent APE approaches utilize transfer learning by adapting pre-trained language or translation models to perform APE~\cite{lopes-etal-2019-unbabels, wei-etal-2020-hw, sharma-etal-2021-adapting}. Also, the recent approaches use multilingual or cross-lingual models to get latent representations of the source and target sentences~\cite{lee-etal-2020-postech}.~\citet{oh-etal-2021-netmarble} have shown that gradually adapting pre-trained models to APE by using the Curriculum Training Strategy (CTS) improves performance.~\citet{deoghare-bhattacharyya-2022-iit} showed that augmenting the APE data with phrase-level APE triplets improves feature diversity, and using a QE system allows for identification and discarding poor-quality APE outputs. We use the APE system to rectify errors in the target side of the noisy pseudo-parallel corpus.

\begin{figure*}
\begin{center}
\includegraphics[width=\textwidth]{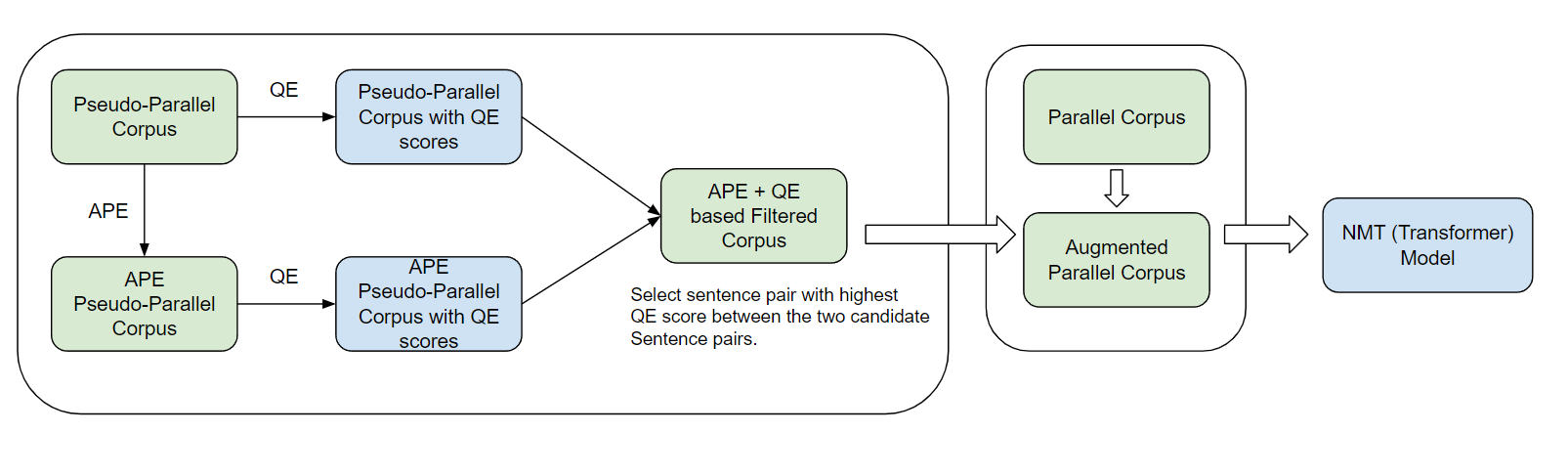}
\caption{The proposed APE and QE assisted corpus filtering pipeline.}
\label{img:ape qe dia}
\end{center}
\end{figure*}

\section{Approaches}
\subsection{LaBSE based Filtering}
Language Agnostic BERT Sentence Embedding model~\cite{labse} is a multilingual embedding model that supports 109 languages, including some Indic languages. We generate the sentence embeddings for the source and target sides of the pseudo-parallel corpora using the LaBSE \footnote{\url{https://huggingface.co/sentence-transformers/LaBSE}} model.
Then, we compute the cosine similarity between the source and target sentence embeddings. After that, we extract good-quality parallel sentences based on a threshold value of the similarity scores.
\subsection{Phrase Pair Injection (PPI) with LaBSE-based Filtering}
~\citet{akshay} proposed a combination of Phrase Pair Injection~\cite{sen2021neural} and LaBSE-based Corpus Filtering to extract high-quality parallel data from a noisy parallel corpus. We train a PBSMT model on the noisy pseudo-parallel corpus using the \textit{Moses} \footnote{\url{http://www2.statmt.org/moses/?n=Development.GetStarted}} decoder. Then, we extract phrase pairs with the highest translation probability. Finally, we perform LaBSE-based filtering on these phrase pairs to remove poor-quality phrase pairs. We augment these high-quality phrase pairs with LaBSE-filtered parallel sentences. 
\subsection{Quality Estimation (QE) based Filtering}
In this approach, we train MonoTransQuest\footnote{\url{https://github.com/TharinduDR/TransQuest}}~\cite{ranasinghe-etal-2020-transquest} model and use it to generate quality scores for the pseudo-parallel corpus of the corresponding language pair. Then, we extract high-quality parallel sentences from the pseudo-parallel corpus using a threshold quality score value.
\subsection{Our Approach, APE + QE based Filtering}
\citet{deoghare-bhattacharyya-2022-iit} proposed a curriculum training strategy to train the APE system. Initially, we use pseudo-parallel corpus (Samanantar~\cite{samanantar}), Anuvaad\footnote{\href{https://github.com/project-anuvaad/anuvaad-parallel-corpus}{Anuvaad: Github Repo}} and ILCI~\cite{bansal2013corpora} and tatoeba\footnote{\href{https://tatoeba.org/en}{Tatoeba Project}} corpus to train an encoder-decoder model to translate English into Marathi. Then, we add another encoder to the model and train the resulting dual-encoder single-decoder model for the APE task. This involves training the model through multiple stages using synthetic APE data and fine-tuning it with real APE data.\\ 
\indent We use the trained APE system to correct errors in the target side of the noisy pseudo-parallel corpus. As APE systems are prone to the problem of ‘over-correction’, we use a sentence-level quality estimation (QE) system to select the final output between an original target sentence and the corresponding output generated by the APE model. 
\section{Experimental Setup}
\label{experimental setup}



\begin{table}[htp]
\centering
\begin{adjustbox}{width=\columnwidth,center}
\begin{tabular}{llr}
\hline
\textbf{Corpus Name} & \textbf{\# Samples} &\textbf{Type}\\
\hline
\textbf{APE data}&18K triplets& Real\\ \hline

 \textbf{Parallel Corpus}&6M pairs&\multirow{3}{*}{Synthetic}\\ 
 \textbf{Phrase-Level Data} & 60K triplets&\\
 \textbf{APE data}& 2.5M triplets\\ \hline
\end{tabular}
\end{adjustbox}
\caption{Dataset Statistics for the task of Automatic Post Editing for En-Mr language pair}
\label{tab:apedata}
\end{table}

\begin{table*}[htp]
\centering
\resizebox{\textwidth}{!}{%
\begin{tabular}{lrrr}
\hline
\textbf{Technique} & \textbf{\# Sentence Pairs (in Millions)}& \textbf{En$\rightarrow$Mr} & \textbf{Mr$\rightarrow$En}\\ \hline
\textbf{APE + QE based Filtering} & 3.5M & \textbf{14.44} & \textbf{25.81}\\
\textbf{APE based Filtering} & 3.5M &14.28&  24.43\\
 \textbf{QE based Filtering}& 2.61M &9.4 & 17.7\\
 \textbf{LaBSE + PPI-LaBSE  based Filtering} & 4.09M& 9.9 & 17.0 \\
 \textbf{LaBSE based Filtering} & 2.85M &  8.8 & 16.7\\
 \textbf{Baseline}& 3.5M& 8.8 & 15.9\\ \hline
\end{tabular}
}
\caption{BLEU scores of NMT models on FLORES101 test data. }
\label{tab:4}
\end{table*}

\subsection{Dataset}
In all NMT experiments, we use two sets of corpus, namely, Parallel and Pseudo-Parallel corpus. The En-Mr \textbf{Parallel corpus} consists of 248K high-quality sentence pairs, which include the ILCI phase 1, Bible, PIB, and PM-India corpus~\cite{ilci, bible, pmi}. The En-Mr \textbf{Pseudo-Parallel} corpus contains 3.28M sentence pairs of varying quality from the Samanantar Corpus~\cite{samanantar}. \\
\indent In QE experiments, we use the QE data used by~\citet{deoghare-bhattacharyya:2022:WMT} for the En-Mr language pair. The En-Mr QE data consists of human-annotated z-standardized Direct Assessment (DA) scores. The train, dev and test instances of QE data are 26K, 1K and 1K, respectively.\\
\indent We use the WMT22 En-Mr APE shared task data to develop the APE system~\cite{bhattacharyya-etal-2022-findings}. The statistics are listed in Table~\ref{tab:apedata}. The train set consists of 18K APE triples generated with the help of professional post-editors. We also use synthetically generated 2.5M APE triplets. The NMT model, which we train as a step in the CTS, is trained using the publically available parallel corpora (Samanantar~\cite{samanantar}, Anuvaad, Tatoeba, and ILCI~\cite{bansal2013corpora}) containing around 6M sentence pairs. The detailed APE data statistics are mentioned in table \ref{tab:apedata}. The annotation guidelines for the task of QE and APE are mentioned in the \textbf{Appendix} \ref{annotation}.

\indent For evaluation, we use the FLORES 101 test set, which contains 1,012 sentence pairs for the En-Mr language pair.

\subsection{Models}
\label{models}
We use a Transformer based architecture to train the NMT models for all our experiments. We use MonoTransQuest model architecture to train the QE models. The training details and model architecture is mentioned in \textbf{Appendix} \ref{app:model arch} and \ref{app:training details}.\\
\noindent \textbf{Baseline:} We train the baseline NMT models on the whole pseudo-parallel corpus augmented with the parallel corpus for the En-Mr language pair.\\
\noindent \textbf{LaBSE based Filtering:} In this model, we use the LaBSE filtering with threshold 0.8 to extract good quality parallel sentences from the En-Mr pseudo-parallel corpus. Then, we augment the parallel corpus with the LaBSE-filtered parallel sentences and train the respective NMT models. \\
\noindent \textbf{LaBSE + PPI-LaBSE based Filtering:} We perform Phrase Pair Injection with LaBSE-based Filtering to train the respective NMT models. This involves extracting LaBSE-filtered parallel sentences and phrases from the pseudo-parallel corpus and augmenting them with the parallel corpora. We train this model for the purpose of comparing it with our proposed best model.\\
\noindent \textbf{Our Model, QE based Filtering:} We train the sentence-level QE model for the En-Mr language pair. We compute the noisy pseudo-parallel corpora quality scores using the trained QE models. Then, we extract high-quality sentence pairs from the pseudo-parallel corpus using the threshold value of -0.5 for the En-Mr language pair. We augment the extracted high-quality sentence pairs with the parallel corpus and train the respective NMT models.\\
\noindent \textbf{Our Model, APE based Filtering:} We train an APE system for the En-Mr language pair and use it to correct errors in the target side of a noisy pseudo-parallel corpus. We augment the corrected pseudo-parallel corpus with the parallel corpus and use it to train the NMT models. \\  
\noindent \textbf{Our Model, APE + QE based Filtering:} First, we train an APE system for the En-Mr language pair. This APE system is then used to correct errors in the target side of a noisy pseudo-parallel corpus. Next, we use a sentence-level QE system to choose the final target sentence, either the original or the corresponding output generated by the APE model. Finally, we augment the corrected pseudo-parallel corpus with the parallel corpus and use it to train the NMT models. 
\section{Results and Analysis}
We evaluate our NMT models using BLEU~\cite{papineni-etal-2002-bleu}. We use SacreBLEU~\cite{sacrebleu} python library to calculate the BLEU scores. Table \ref{tab:4} shows that \textbf{APE + QE based filtering} model outperforms all other models for the En-Mr language pair. The \textbf{APE + QE based filtering} model improves the MT system's performance by \textbf{5.64} and \textbf{9.91} BLEU points over the \textbf{baseline} model for En$\rightarrow$Mr and Mr$\rightarrow$En, respectively. It also outperforms \textbf{QE based Filtering} and \textbf{LaBSE + PPI-LaBSE based Filtering} model by \textbf{5.04}, \textbf{8.11} BLEU points for En$\rightarrow$Mr and \textbf{4.54}, \textbf{8.81} BLEU points for Mr$\rightarrow$En, respectively.\\
\indent We observe a significant improvement in the En-Mr NMT system by \textbf{5.48} and \textbf{8.53} BLEU points over the baseline, merely by correcting errors in the target side of the noisy pseudo-parallel corpus using the En-Mr APE system. Furthermore, we have also discovered that utilizing QE-based sentence selection alongside APE-based pre-editing of the pseudo-parallel corpus resulted in a further enhancement of the En-Mr NMT system by \textbf{0.16} and \textbf{1.38} BLEU points when compared to the \textbf{APE-based filtering model}. 


\section{Conclusion and Future Work}
In this work, we show that correcting errors in the target side of the noisy pseudo-parallel corpus using an APE model and selecting high-quality sentence pairs from the original and corrected sentence pairs using QE helps NMT models improve their performance significantly.
Also, the proposed approach is independent of the properties of English and Marathi languages and can be applied to any language pair. However, our language pair-agnostic approach requires APE and QE data to correct and filter noise from pseudo-parallel data. \\
\indent In the future, we will use the proposed approach and analyze data from other language pairs. We also plan to use a combination of word and sentence level QE to select high-quality sentence pairs between original and APE-corrected sentence pairs.

\section*{Limitations}
Developing APE and QE systems involves using human-annotated data, which can be costly and time-consuming. We restrict our experiments solely to the En-Mr language pair due to a lack of APE and QE resources. The QE-based filtering experiments involve a hyper-parameter called ``threshold quality score." We conduct experiments with different values of this hyper-parameter to achieve optimal results. 

\section*{Ethics Statement}
The aim of our work is to improve the quality of En-Mr MT systems by using APE and QE models to rectify errors on the target side of the pseudo-parallel corpus and extract high-quality parallel corpus. The datasets that we used in this work are publicly available, and we have cited the sources of all the datasets that we have used.  Publicly available datasets can contain biased sentences.



\bibliography{anthology,custom}
\bibliographystyle{acl_natbib}

\appendix

\section{Appendix}
\label{sec:appendix}

\subsection{Model Architecture (Referred from section \ref{models}, line no. 233)}
\label{app:model arch}
We use a Transformer based architecture to train English-Marathi NMT models for all our experiments. The encoder of the Transformer consists of 6 encoder layers and 8 encoder attention heads. The encoder uses embeddings of dimension 512. The decoder of the Transformer also consists of 6 decoder layers and 8 decoder attention heads. The English-Marathi NMT models consist of around 66M parameters. We use MonoTransQuest architecture to train the English-Marathi QE model for all our experiments. A transformer-based two-encoder single-decoder architecture having around 40M parameters is used for developing the APE system. The encoders use the same architecture as IndicBERT and are also initialized using the IndicBERT weights. The decoder contains two cross-attention layers with 8 attention heads. The size of the APE model
\renewcommand{\figurename}{Figure}
\renewcommand{\thefigure}{2}
\begin{figure}[htp]

\begin{center}
\includegraphics[scale=0.5]{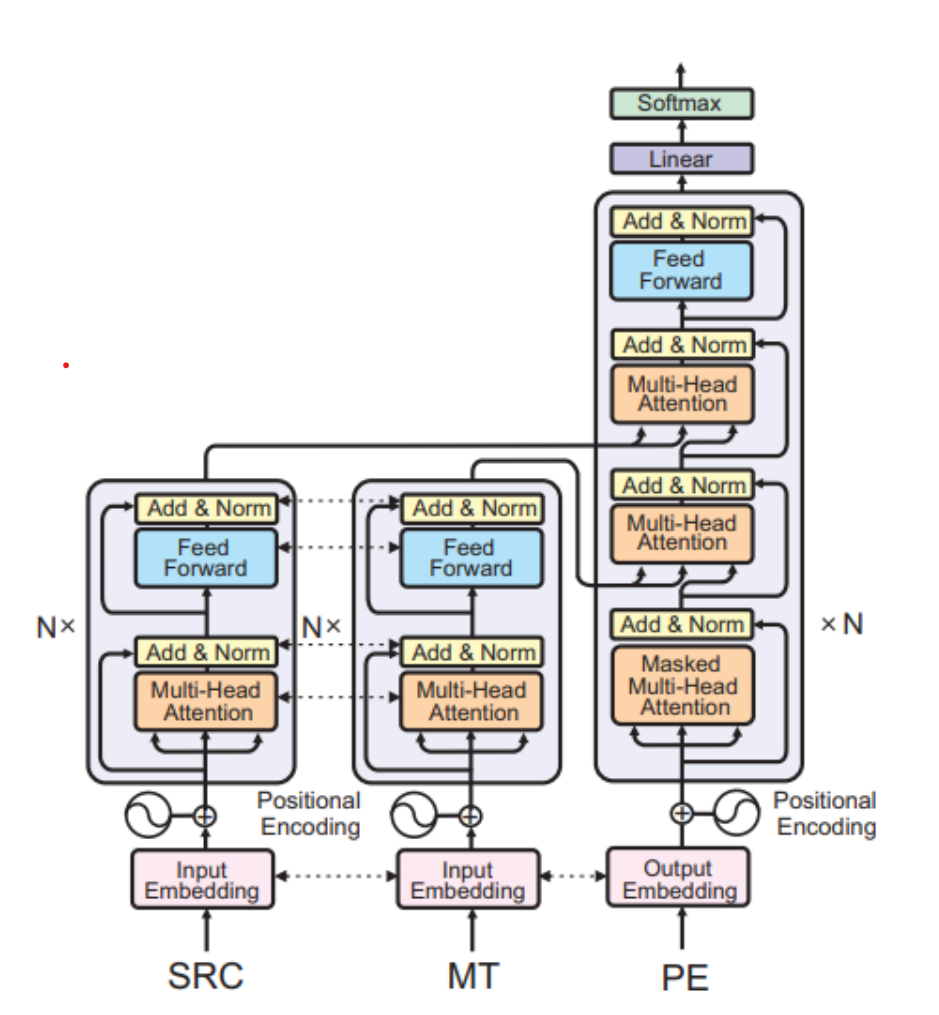}\caption{ Dual-encoder Single Decoder Architecture. Dashed arrows represent tied parameters and common embedding matrices for encoders and decoder~\cite{deoghare-bhattacharyya-2022-iit}}
\label{ape:arch}
\end{center}
\end{figure}

\subsection{Training Details (Referred from section \ref{models}, line no. 233)}
\label{app:training details}
We used the Indic NLP library for preprocessing the Indic language data and  Moses for preprocessing the English language data. For Indic languages, we normalize and tokenize the data. For English, we lowercase and tokenize the data.
We use the OpenNMT-py library to train the Transformer based NMT models. The hyperparameter values are selected using manual tuning. The optimizer used was adam with betas (0.9, 0.98). The initial learning rate used was 5e-4 with the inverse square root learning rate scheduler. We use 8000 warmup updates. The dropout probability value used was 0.1 and the criterion used was label smoothed cross entropy with label smoothing of 0.1. We use a batch size of 4096 tokens. All the models were trained for 200,000 training steps. We use MonoTransquest\footnote{\url{https://github.com/TharinduDR/TransQuest}} model to train the sentence-level QE model. We start with a learning rate of 2e-5 and use 5\% of training data for warm-up. We use early patience over ten steps. We use a batch size of eight. 
For training the APE system, we use a batch size of 16. We set the number of epochs to 1000 and use the early stopping to halt the training by using patience over the 20 training steps. We use the Adam optimizer with an initial learning rate of 5e-5 and the 0.9 and 0.997 beta values. Both encoders of the model are initialized using IndicBERT weights. 
We use a single Nvidia A100 GPU with 40 GB memory to train our NMT APE, and QE models.
Training the APE model using CTS and including the fine-tuning, takes around 48 hours when using a single GPU.

\subsection{Annotation Details (Referred from section \ref{experimental setup}, line no. 224)}
\label{annotation}

\subsubsection{QE Guidelines}
The guidelines provided to the annotators for the Quality Estimation task are shown in Figure \ref{fig:i6}.
\renewcommand{\figurename}{Figure}
\renewcommand{\thefigure}{3}
\begin{figure*}[htp]

\begin{center}
\includegraphics[scale=0.9]{Images/qe task.png}\caption{Evaluation Guidelines for the English-Marathi Direct Assessment (for Quality Estimation) Task.  }
\label{fig:i6}
\end{center}
\end{figure*}
\subsubsection{APE Guidelines}
The guidelines provided to the annotators for the Automatic Post-Editing task are shown in Figure \ref{ape:guide}.
\renewcommand{\figurename}{Figure}
\renewcommand{\thefigure}{4}
\begin{figure*}[htp]

\begin{center}
\includegraphics[scale=0.8]{Images/Post-Editing Guidelines_page-0001.jpg}\caption{Guidelines for the English-Marathi Automatic Post-Editing Task.}
\label{ape:guide}
\end{center}
\end{figure*}
\end{document}